\documentclass[letterpaper, 10 pt, conference]{ieeeconf}  
\usepackage{xcolor}

\IEEEoverridecommandlockouts                              

\usepackage{caption}
\usepackage{graphicx}
\usepackage{graphics} 
\usepackage{epsfig} 
\usepackage{mathptmx} 
\usepackage{times} 
\usepackage{amsmath} 
\usepackage{amssymb}  

\usepackage{booktabs}
 
\usepackage{xspace}
\usepackage{balance}
\usepackage{comment}
\usepackage{gensymb} 
\usepackage{multirow}
\usepackage{bbm} 
\usepackage{subfigure} 

\usepackage{xcolor}

\usepackage{color, colortbl}
\definecolor{lightgray}{gray}{0.85}

\DeclareMathAlphabet{\mathcal}{OMS}{cmsy}{m}{n}

\title{\LARGE \bf
Uncertainty-Aware Vehicle Orientation \\Estimation for Joint Detection-Prediction Models
}

\author{ 
  Henggang Cui, Fang-Chieh Chou, Jake Charland, Carlos Vallespi-Gonzalez, Nemanja Djuric \\
  \thanks{
    Authors are with Uber Advanced Technologies Group (ATG), 50 33rd Street, Pittsburgh, PA 15201; emails:
     {\tt\small \{hcui2, fchou, jakec, cvallespi,  ndjuric\}@uber.com}
  }
}

\begin{document}

\maketitle
\thispagestyle{empty}
\pagestyle{empty}

\begin{abstract}
Object detection is a critical component of a self-driving system, tasked with inferring the current states of the surrounding traffic actors. 
While there exist a number of studies on the problem of inferring the position and shape of vehicle actors, understanding actors' orientation remains a challenge for existing state-of-the-art detectors. 
Orientation is an important property for downstream modules of an autonomous system, particularly relevant for motion prediction of stationary or reversing actors where current approaches struggle. 
We focus on this task and present a method that extends the existing models that perform joint object detection and motion prediction, allowing us to more accurately infer vehicle orientations.
In addition, the approach is able to quantify prediction uncertainty, outputting the probability that the inferred orientation is flipped, which allows for improved motion prediction and safer autonomous operations.
Empirical results show the benefits of the approach, obtaining state-of-the-art performance on the open-sourced nuScenes data set.

\end{abstract}

\section{Introduction}
\label{sect:introduction}

In order to operate safely and efficiently in a real-world, a self-driving vehicle (SDV) needs to be able to infer the current state of its surroundings, as well as to predict how this state would change in the near future. 
This task is addressed by object detection and motion prediction modules, two critical components of an autonomous system \cite{urmson2008self,casas2018intentnet,lasernet}.
The traditional approach to implementing these modules is in a sequential manner \cite{urmson2008self}, where the two models are trained and run separately.
In particular, a detection model processes raw sensor inputs to infer object detections and their states (such as position, dimensions, and velocity) \cite{casas2018intentnet}, which are in turn used as an input to a motion prediction model that outputs objects' future trajectories as well as the uncertainty of their motion \cite{djuric2020wacv}.
Going beyond the sequential approach, researchers recently proposed to combine the two components into a unified, end-to-end model, shown to achieve exemplary performance on both tasks \cite{casas2018intentnet,djuric2020multixnet}. 
These state-of-the-art models are the focus of our current work.

Due to its importance for SDV operations, the task of object detection has sparked a lot of interest from the research community \cite{guo2020deep}, leading to significant improvements in the models' performance over the past years.
However, while achieving strong results, when it comes to vehicle detections most of the recent research was focused on estimating a limited set of object states, namely positions and bounding boxes. 
Nevertheless, object orientation (defined as a direction of the front of the vehicle) is an important information that the autonomous system requires to improve safety and efficiency, allowing better and more accurate future motion prediction. 
Moreover, an important aspect of orientation estimation is the modeling of its uncertainty \cite{mousavian20173d}, providing a more complete view of the stochastic SDV environment. 
Understanding the full orientation state is particularly critical in the case of vehicles that are static, slow-moving, or reversing, where behavior uncertainty is high and common heuristic \cite{cui2020deep} that computes actor orientation from the inferred travel direction of future trajectories breaks down.
More involved methods are needed in order to address these issues, and although the problem of better orientation prediction has received some attention previously, the accuracy of the current state-of-the-art models remains suboptimal.

\begin{table*} [t!]
\centering
\caption{Overview of various orientation estimation methods}
\label{tab:all-methods}
{\normalsize
{
  \begin{tabular}{llcll}
    {\bf Method} & {\bf References} & {\bf Full-range} & {\bf Model output} & {\bf Loss} \\
    \hline
    \rowcolor{lightgray}
    \texttt{Sin-cos-2x} & \cite{djuric2020multixnet, yang2018hdnet} &  & $\sin(2 \hat \theta)$, $\cos(2 \hat \theta)$ & $\ell_1\big({\sin}(2 \hat \theta) - \sin(2 \theta)\big) + \ell_1\big({\cos}(2 \hat \theta) - \cos(2 \theta)\big)$ \\
    \texttt{L1-sin}&\cite{yan2018second, lang2019pointpillars, zhou2020end} &  & $\hat \theta$ & $\ell_1\big({\sin}(\hat \theta - \theta)\big)$ \\
    \rowcolor{lightgray}
    \texttt{Sin-cos}&\cite{luo2018fast} & $\checkmark$ & $\sin(\hat \theta)$, $\cos(\hat \theta)$ & $\ell_1\big({\sin}(\hat \theta) - \sin(\theta)\big) + \ell_1\big({\cos}(\hat \theta_t) - \cos(\theta)\big)$ \\
    \texttt{MultiBin}&\cite{mousavian20173d} & $\checkmark$ & $\{\Delta \hat \theta_i, \hat p_i\}_{i=1}^n$ & $\mathcal{L}_\text{multibin}$~\cite{mousavian20173d} \\
    \rowcolor{lightgray}
    \texttt{Flip-aware} & ours & $\checkmark$ & $\sin(\hat \theta)$, $\cos(\hat \theta)$, $\hat p_f$ & $\mathcal{L}_\text{final}$ from equation~\eqref{eq:flipping_aware_loss} \\
    \hline
\end{tabular}
}
}
\end{table*}

The existing work on orientation estimation of detection bounding boxes can be split into two categories of methods.
The first category, which we refer to as \emph{full-range} methods, estimates the orientations in the $(-180\degree, 180\degree]$ range~\cite{luo2018fast, mousavian20173d}, allowing estimation of the exact actor state.
The second category, which we refer to as \emph{half-range} methods, only estimates the vehicle orientations in the limited $(-90\degree, 90\degree]$ range~\cite{djuric2020multixnet, yang2018hdnet, yan2018second, lang2019pointpillars, zhou2020end}, and can be used in applications where a full understanding of object orientation is not that relevant (such as in pure detection applications).
It was shown that the half-range methods achieve better detection performance than the full-range methods, however, by design, they are not able to distinguish the front and back of a bounding box, which is important for the SDV system and its motion prediction task.
In the experimental section, we revisit these two types of methods and provide a detailed evaluation of the pros and cons of both approaches.

In this paper, we take the above-mentioned considerations into account and propose a novel uncertainty-aware full-range method for orientation estimation.
The approach is able to achieve detection performance that is comparable to or better than the half-range methods while being able to estimate the actor's full-range orientation as well as the probability that the estimation is flipped from the ground-truth by $180 \deg$.
We summarize our contributions below:
\begin{itemize}
\item we study the trade-off between full-range orientation estimation capability and detection accuracy using the existing state-of-the-art methods;
\item we describe a novel method to estimate full-range orientations without losing detection accuracy;
\item the proposed method is able to quantify the flipped uncertainty in that estimate.
\end{itemize}

\section{Related work}
\label{sec:related-work}

\subsection{Object detection for autonomous driving}
Detecting objects in SDV's surroundings is a critical task of autonomous systems, required for safe road operations.
Lidar-based approaches have proven to be the workhorse within the research community \cite{guo2020deep}, with a large number of methods proposed recently to address the detection task.
Deep neural networks are the most popular choice shown to achieve state-of-the-art performance, and various deep methods differ by how the lidar data is presented to the model. 
The authors of \cite{yang2018pixor} proposed to encode the lidar points into a bird's-eye view (BEV) grid, and following a number of convolutional layers directly regress probability existence and object state for each grid cell.
If available, additional information can be fused in the BEV grid in a straightforward manner, such as a high-definition map \cite{pmlr-v87-yang18b} or radar data \cite{yang2020radarnet}.
An alternative approach is to project the inputs into the range view (RV) grid, encoding the sensor data in lidar's native representation \cite{meyer2019lasernet}.
The authors of \cite{Meyer_2019_CVPR_Workshops} extended the RV-based method with fused camera data, which is natively captured in a front-view frame.

Recently, detection methods were extended to also solve the prediction task, giving rise to unified, end-to-end approaches operating in the BEV frame \cite{luo2018fast}. 
Authors of \cite{casas2018intentnet} proposed to also infer detected actor's high-level intents, while authors of MultiXNet \cite{djuric2020multixnet} introduced a two-stage architecture that achieved state-of-the-art performance in both detection and prediction metrics.
In our current work we take MultiXNet as the baseline, and propose a loss formulation that leads to significantly improved orientation accuracy.

\subsection{Orientation estimation methods}

The \emph{full-range} orientation estimation methods attempt to estimate actors' orientations in the full $(-180\degree, 180\degree]$ range, as exemplified by the \texttt{Sin-cos} approach~\cite{luo2018fast}.
This method maps each orientation value $\theta$ into two independent targets, $\sin(\theta)$ and $\cos(\theta)$, which are independently trained using two smooth-L1 loss terms, denoted as $\ell_1$.
Then, during inference, the final orientation can be computed as $\arctan(\sin(\theta), \cos(\theta))$, although note that the two output values are not guaranteed to be normalized.
However, with this setup, a flipped orientation prediction (i.e., $180 \degree$ orientation error) is penalized by a large loss and causes the two target outputs to move across the unit circle instead of around.
Indeed, our experiments indicate that this characteristic of the loss hurts the overall model performance, leading to suboptimal detection and prediction accuracies.
Moreover, as in this paper, we consider a LiDAR-based model, estimating the front and back of a bounding box is a challenging problem for such systems because the front and back of a vehicle may not be easily distinguishable from the LiDAR point cloud, especially for objects that are far away from the sensor with a few LiDAR returns.

To address this issue, several detection models took a step back and proposed the \emph{half-range} methods that estimate orientation in the limited $(-90\degree, 90\degree]$ range.
Examples of such orientation estimation methods include \texttt{Sin-cos-2x} and \texttt{L1-sin} approaches.
MultiXNet~\cite{djuric2020multixnet} and HDNet~\cite{yang2018hdnet} used the \texttt{Sin-cos-2x} method that represents each orientation value as $\sin(2 \theta)$ and $\cos(2 \theta)$.
Similarly to above-mentioned \texttt{Sin-cos}, the two targets are trained independently with the $\ell_1$ loss, and the final orientation is computed as $0.5\arctan(\sin(2 \theta), \cos(2 \theta))$.
On the other hand, \mbox{SECOND~\cite{yan2018second}}, PointPillars~\cite{lang2019pointpillars}, and Zhou et al.~\cite{zhou2020end} used the \texttt{L1-sin} method that directly regresses the orientation values trained using the loss $\ell_1(\sin(\theta))$.
As we show in our evaluation results, the half-range methods achieve better detection accuracies than full-range methods in terms of average precision, yet by design, they are not able to accurately estimate the front and back of the vehicles, which is critical for the autonomous driving task.
Table~\ref{tab:all-methods} summarizes the discussed state-of-the-art methods for orientation estimation.

\subsection{Uncertainty-aware orientation estimation}

Beyond predicting the orientation itself, understanding its uncertainty is another important task that allows for safer autonomous operations.
Mousavian et al.~\cite{mousavian20173d} proposed the MultiBin method for multimodal full-range orientation estimation.
They proposed to bin the orientations into $n$ overlapping bins, and for each bin to output two values, a probability that the object orientation lies within the bin and the residual angle correction relative to the bin's central angle.
They trained the bin probabilities with the cross-entropy loss and the residual angle corrections of the matching bins with the cosine distance loss.
In such a way, the model can produce multiple orientation estimations for an actor, along with their probabilities.
The method, however, requires one to tune the number of bins as well as their placement, whereas our method does not require any extra hyper-parameters.
Moreover, our work estimates a full-range orientation along with its uncertainty without the discretization step, thus simplifying the learning problem.
The experimental results will show that our proposed method outperforms the state-of-the-art MultiBin method.



\section{Methodology}

While the proposed approach is generic and can be applied to any model architecture, we implemented and evaluated it on top of MultiXNet~\cite{djuric2020multixnet}, a state-of-the-art joint object detection and motion prediction model that employs a two-stage architecture.
The model takes as input a total of $T$ current and historical LiDAR sweeps along with a high-definition map of SDV's surroundings, which are rasterized onto a BEV grid. 
The method then applies a multi-scale convolutional network on the resulting raster, outputting existence probability and bounding box for each grid cell which completes the first stage.
As a part of the second stage, the feature maps corresponding to detected objects are cropped and further processed by a sequence of convolutional layers, eventually outputting each actor's future trajectories for a total of $H$ time steps.
The actors' bounding boxes are parameterized by center position, width and height dimensions, and orientation, where for the orientation the model predicts the yaw component for all future time steps, denoted as $\{\hat{\theta}_t\}_{t=1}^H$.
For more detailed discussion we refer the reader to \cite{djuric2020multixnet}, omitted here due to space limitations.

The original MultiXNet model uses the \texttt{Sin-cos-2x} orientation estimation method that represents the orientations as $2 H$ independent targets $\{\sin(2 \hat \theta_t), \cos(2 \hat \theta_t)\}_{t=1}^H$, which are trained with a \emph{half-range loss} given as
\begin{align}
\begin{split}
\label{eq:half_range_loss}
    \mathcal{L}_\text{half} = \sum_{t=1}^H &\ell_1\big(\sin(2 \hat \theta_t) - \sin(2 \theta_t)\big) \\
    &+ \ell_1\big(\cos(2 \hat \theta_t) - \cos(2 \theta_t)\big),
\end{split}
\end{align}
where $\ell_1$ denotes smooth-L1 loss, and $\{\theta_t\}_{t=1}^H$ are ground-truth labels.
The $\sin(2 \theta_t)$ and $\cos(2 \theta_t)$ targets are not guaranteed to be normalized, and the final orientation can be computed during inference as $\theta_t = 0.5\arctan \big(\sin(2 \theta_t), \cos(2 \theta_t)\big)$, given in the $(-90\degree, 90\degree]$ range.

\subsection{Combining half-range and full-range losses}
\label{sect:half_full_combo}

In order to estimate the full range orientations,
our proposed method represents the orientations as $\{\sin(\hat \theta_t), \cos(\hat \theta_t)\}_{t=1}^H$, and similarly to the \texttt{Sin-cos} method~\cite{luo2018fast} we define a \emph{full-range loss} computed as
\begin{align}
\begin{split}
\label{eq:full_range_loss}
    \mathcal{L}_\text{full} = \sum_{t=1}^H &\ell_1\big(\sin(\hat \theta_t) - \sin(\theta_t)\big) \\
    &+ \ell_1\big(\cos(\hat \theta_t) - \cos(\theta_t)\big).
\end{split}
\end{align}

To bridge the detection performance gap between full-range and half-range orientation estimation methods, we propose to extend~\eqref{eq:full_range_loss} with an additional half-range loss term from~\eqref{eq:half_range_loss},
where the half-range representation parameters in~\eqref{eq:half_range_loss} can be computed from the full-range parameters using the following trigonometric identities,
\begin{align}
\begin{split}
\label{eq:convert_sin2cos2}
    \sin(2 \hat\theta_t) &= 2 \sin(\hat \theta_t) \cos(\hat \theta_t), \\
    \cos(2 \hat \theta_t) &= {\cos^2}(\hat \theta_t) - {\sin^2}(\hat \theta_t).
\end{split}
\end{align}
As will be shown in the evaluation results presented in Section~\ref{sec:quantitative_results}, combining the losses leads to significant improvements of the model performance.

\subsection{Flipping-aware orientation prediction}
\label{sect:flipping_aware}

Even with the combined loss, the model may still incur a high penalty from the $\mathcal{L}_\text{full}$ component when outputting a bounding box orientation that is flipped by $180\degree$.
This is a particular problem for static objects, where considering historical LiDAR sweeps provides little evidence of their orientation.
To mitigate this problem, we propose a novel formulation that renders the full-range loss \emph{flipping-aware}.

In particular, in addition to full-range orientation outputs $\{\sin(\hat \theta_t), \cos(\hat \theta_t)\}_{t=1}^H$, the model is trained to also predict a probability that the orientation is flipped by $180\degree$, denoted by $\hat{p}_f$.
In other words, $\hat{p}_f$ is an indicator of whether $\arctan(\sin(\hat \theta_t), \cos(\hat \theta_t))$ or $\arctan(-\sin(\hat \theta_t), -\cos(\hat \theta_t))$ is the true orientation of the bounding box.

In order to train such a model, we define a \emph{flipped full-range loss} $\mathcal{L}_\text{flipped}$ as given here,
\begin{align}
\begin{split}
\label{eq:flipped_full_range_loss}
    \mathcal{L}_\text{flipped} = \sum_{t=1}^H &\ell_1\big(\sin(\hat \theta_t + 180\degree) - \sin(\theta_t)\big) \\
    &+ \ell_1\big(\cos(\hat \theta_t + 180\degree) - \cos(\theta_t)\big) \\
    = \sum_{t=1}^H &\ell_1\big(-\sin(\hat \theta_t) - \sin(\theta_t)\big) \\
    &+ \ell_1\big(-\cos(\hat \theta_t) - \cos(\theta_t)\big).
\end{split}
\end{align}
Then, for each actor we compare the values of $\mathcal{L}_\text{full}$ and $\mathcal{L}_\text{flipped}$ losses and use $p_f = \mathbbm{1}_{\mathcal{L}_\text{full} > \mathcal{L}_\text{flipped}}$ as the ground-truth classification label to train the flipped classification output $\hat{p}_f$, where $\mathbbm{1}_{cond}$ is an indicator function equaling 1 if condition $cond$ is true and 0 otherwise.
Then, we define our final loss as follows,
\begin{align}
\begin{split}
\label{eq:flipping_aware_loss}
    \mathcal{L}_\text{final} = &\mathcal{L}_\text{half}+ \min(\mathcal{L}_\text{full}, \mathcal{L}_\text{flipped}) \\
    &+ \text{CrossEntropy}(\hat{p}_f, \mathbbm{1}_{\mathcal{L}_\text{full} > \mathcal{L}_\text{flipped}}).
\end{split}
\end{align}

Note that the loss penalizes only the minimum of $\mathcal{L}_\text{full}$ and $\mathcal{L}_\text{flipped}$.
Thus, if the model predicts a flipped orientation for the bounding box, it will not be penalized by the orientation loss but by the flipped classification loss instead.
As a result, when the model predicts a flipped orientation it will be encouraged to keep pushing the orientation closer to the flipped ground-truth orientation while moving the flipped probability closer to $1$.
Since the model only needs to predict one extra orientation flipped probability value for each actor, our method has a neglectable impact on the model inference speed.
Lastly, following the training completion we implement a post-processing step to flip the orientations whose probabilities are greater than $0.5$, where we also update their flipped probabilities as $\hat{p}_f \xleftarrow[]{}1-\hat{p}_f$.


\begin{figure*}[h]
    \centering
    \subfigure{\includegraphics[width=0.27\textwidth]{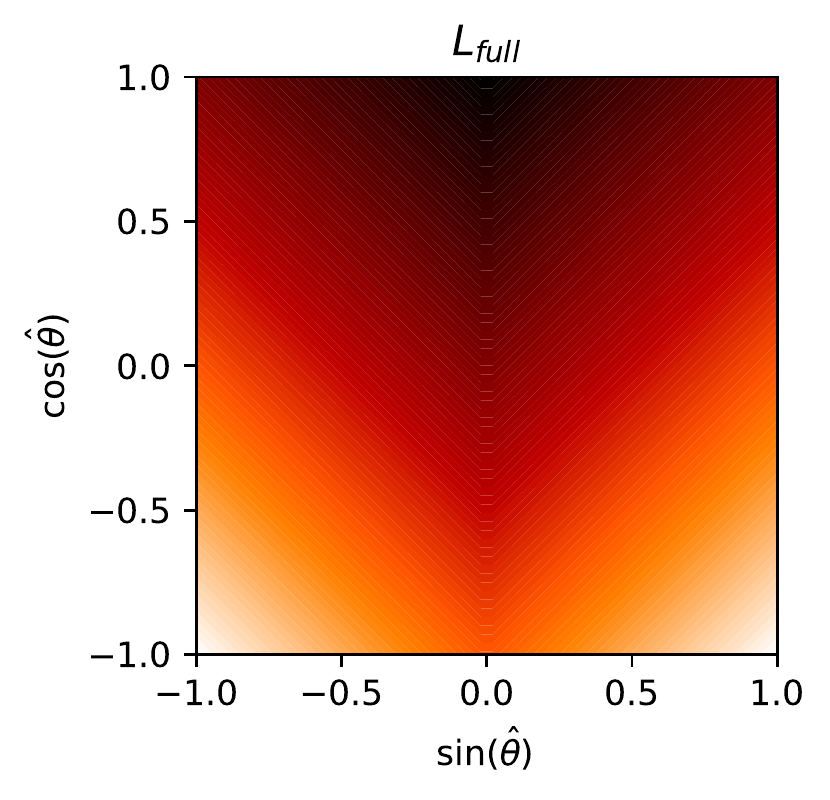}}
    \subfigure{\includegraphics[width=0.27\textwidth]{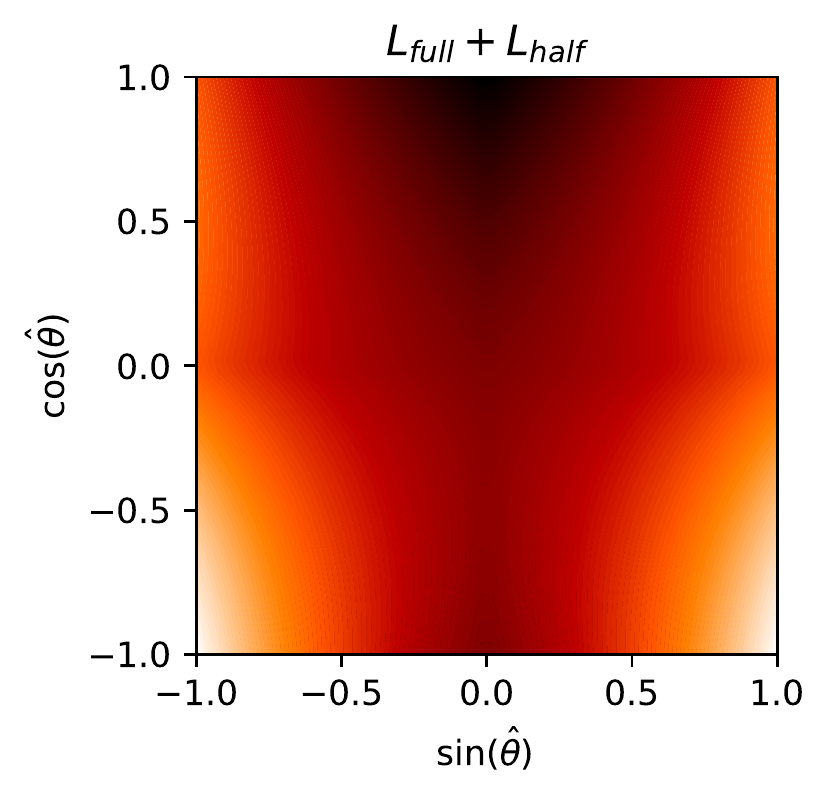}} \\
    \subfigure{\includegraphics[width=0.27\textwidth]{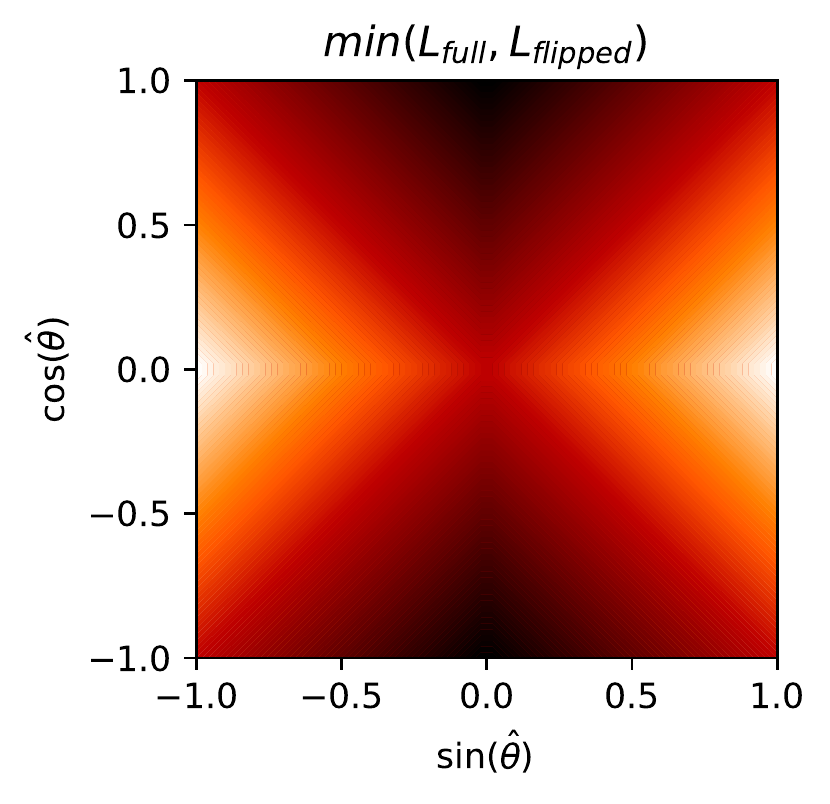}}
    \subfigure{\includegraphics[width=0.27\textwidth]{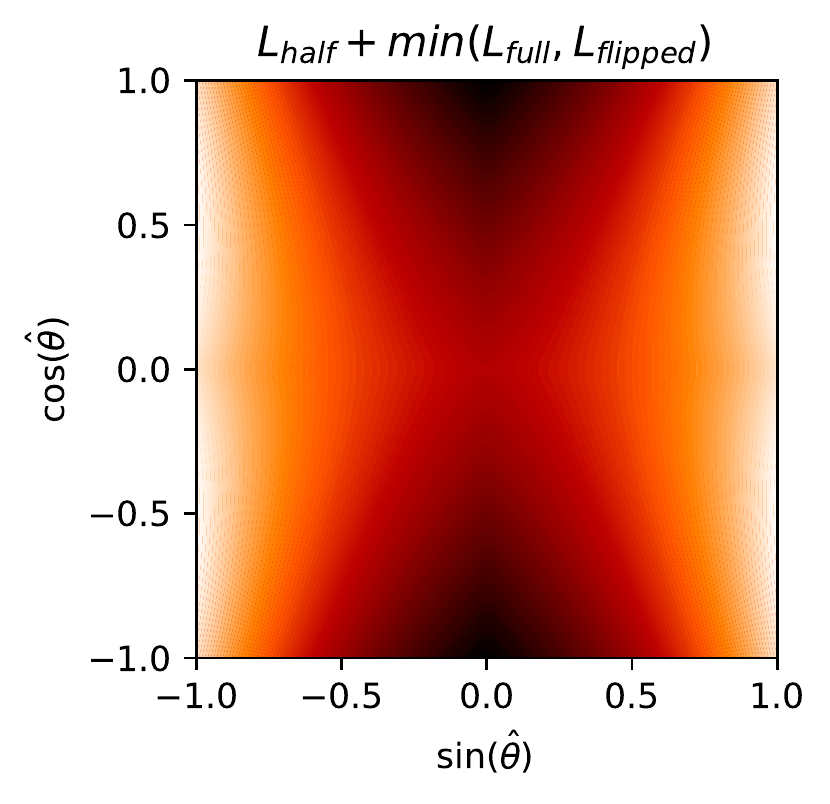}}
    \caption{Visualization of losses, with darker color indicating lower loss and ground truth corresponding to $(\sin \theta, \cos \theta) = (0, 1)$}
    \label{fig:loss_analysis}
    \vspace{-0.1in}
\end{figure*}

\subsection{Loss analysis}

In Figure~\ref{fig:loss_analysis} we provide an analysis of the loss landscapes of various orientation losses considered in this work, with the ground truth corresponding to $(\sin \theta, \cos \theta) = (0, 1)$. For simplicity, the plots do not include the cross-entropy loss term introduced in equation \eqref{eq:flipping_aware_loss}.

In Figure~\ref{fig:loss_analysis}(a) we can see that $\mathcal{L}_\text{full}$ from \eqref{eq:half_range_loss} has only one minimum at the ground-truth point $(0, 1)$, and has a very high loss at the point $(0, -1)$ which corresponds to the $180\degree$-flipped orientation.
On the other hand, in Figure~\ref{fig:loss_analysis}(b) we illustrate $\mathcal{L}_\text{full} + \mathcal{L}_\text{half}$ introduced in Section \ref{sect:half_full_combo}, which mitigates this issue with the addition of the half-range loss.
It has two local minima, the ground truth at $(0, 1)$ and the flipped ground-truth point $(0, -1)$, where $(0, 1)$ is the global minimum. 
Note that orientation error of $180\degree$ at $(0, -1)$ has lower loss than lower-degree errors (e.g., $(1, 0)$ that corresponds to orientation error of $90\degree$), which encourages the model to prioritize estimation of half-range over the full-range orientation in the case of errors larger than $90\degree$.

When it comes to the losses $\min(\mathcal{L}_\text{full}, \mathcal{L}_\text{flipped})$ and $\min(\mathcal{L}_\text{full}, \mathcal{L}_\text{flipped}) + \mathcal{L}_\text{half}$ introduced in Section \ref{sect:flipping_aware} and illustrated in Figure~\ref{fig:loss_analysis}(c) and Figure \ref{fig:loss_analysis}(d), respectively, both have two global minima located at the ground-truth point $(0, 1)$ and the flipped ground-truth point $(0, -1)$.
As a result, they incur no penalty for $180\degree$ errors and instead rely on the cross-entropy term to encourage the model to distinguish the front and back of the bounding box.
\begin{table*} [t!]
\centering
\caption{Quantitative comparison of the competing approaches; confidence intervals are computed over 3 runs}
\label{tab:main_results}
{\normalsize
{
  \begin{tabular}{lccccccc} 
     &  &  & \multicolumn{3}{c}{\bf Orientation error [deg] $\downarrow$} & & \\
      \cmidrule(lr){4-6}
     & & & {\bf Half-range} & \multicolumn{2}{c}{\bf Full-range} & \multicolumn{2}{c}{\bf $\boldsymbol \ell_2$@$3\text{s}$ [m] $\downarrow$} \\
      \cmidrule(lr){5-6} \cmidrule(lr){7-8}
    {\bf Method} & {\bf $\text{AOS}_{0.7}$ $\uparrow$} & {\bf $\text{AP}_{0.7}$ $\uparrow$} & {\bf All} & {\bf All} & {\bf Moving} & {\bf All} & {\bf Moving} \\
    \hline
    \rowcolor{lightgray}
    \texttt{Sin-cos-2x} & 40.7 {\footnotesize $\pm$ 0.7} & {\bf 60.8 {\footnotesize $\pm$ 1.0}} & {\bf 1.72 {\footnotesize $\pm$ 0.04}} & 59.9 {\footnotesize $\pm$ 1.4} & 4.7 {\footnotesize $\pm$ 0.5} & 0.99 {\footnotesize $\pm$ 0.02} & 2.80 {\footnotesize $\pm$ 0.06} \\
    \texttt{L1-sin} & 39.8 {\footnotesize $\pm$ 0.1} & 59.5 {\footnotesize $\pm$ 0.1} & 2.06 {\footnotesize $\pm$ 0.10} & 60.2 {\footnotesize $\pm$ 0.4} & 5.0 {\footnotesize $\pm$ 0.1} & {\bf 0.97 {\footnotesize $\pm$ 0.01}} & 2.76 {\footnotesize $\pm$ 0.03} \\
    \rowcolor{lightgray}
    \texttt{Sin-cos} & 55.1 {\footnotesize $\pm$ 0.5} & 57.1 {\footnotesize $\pm$ 0.5} & 2.32 {\footnotesize $\pm$ 0.01} & {\bf 8.2 {\footnotesize $\pm$ 0.2}} & 2.4 {\footnotesize $\pm$ 0.2} & 1.01 {\footnotesize $\pm$ 0.02} & 2.79 {\footnotesize $\pm$ 0.06} \\
    \texttt{MultiBin-2} & 55.0 {\footnotesize $\pm$ 0.3} & 57.3 {\footnotesize $\pm$ 0.6} & 2.54 {\footnotesize $\pm$ 0.01} & 9.4 {\footnotesize $\pm$ 0.7} & 3.0 {\footnotesize $\pm$ 0.1} & 1.01 {\footnotesize $\pm$ 0.01} & {\bf 2.73 {\footnotesize $\pm$ 0.05}} \\
    \rowcolor{lightgray}
    \texttt{MultiBin-4} & 55.5 {\footnotesize $\pm$ 0.3} & 58.0 {\footnotesize $\pm$ 0.5} & 2.14 {\footnotesize $\pm$ 0.07} & 9.8 {\footnotesize $\pm$ 0.5} & 2.6 {\footnotesize $\pm$ 0.3} & 1.01 {\footnotesize $\pm$ 0.01} & 2.78 {\footnotesize $\pm$ 0.04} \\
    \texttt{Flip-aware} & {\bf 57.9 {\footnotesize $\pm$ 0.4}} & {\bf 60.7 {\footnotesize $\pm$ 0.2}} & {\bf 1.71 {\footnotesize $\pm$ 0.04}} & 9.6 {\footnotesize $\pm$ 0.8} & {\bf 2.2 {\footnotesize $\pm$ 0.1}} & 0.99 {\footnotesize $\pm$ 0.01} & {\bf 2.75 {\footnotesize $\pm$ 0.03}} \\
    \hline
\end{tabular}
}
}
\end{table*}


\begin{figure*}[t!]
    \centering
    \subfigure{
        \label{fig:flipped_prob_vs_foe}
        \includegraphics[width=0.25\textwidth]{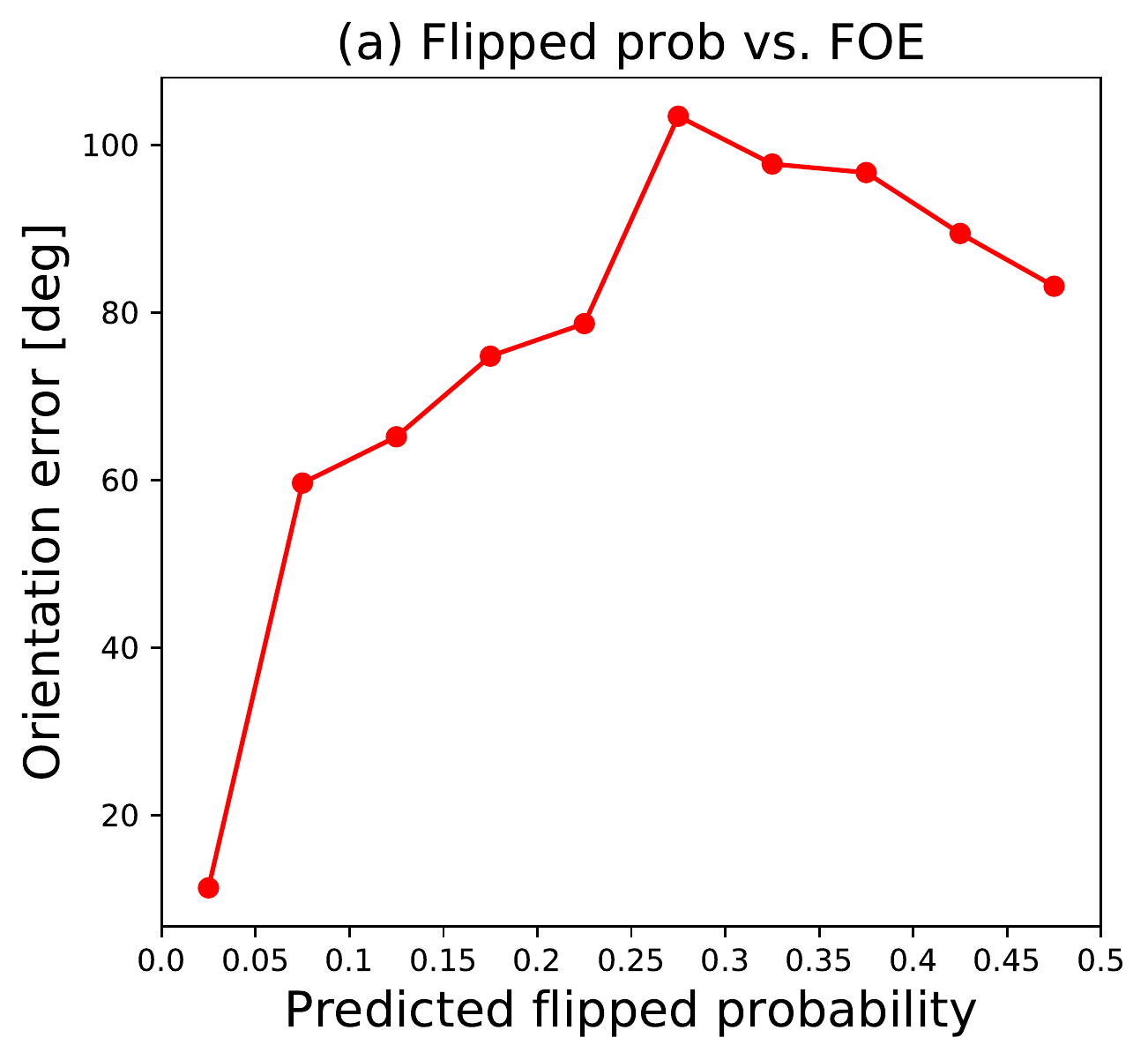}
    }
    \subfigure{
        \label{fig:flipped_prob_vs_speed}
        \includegraphics[width=0.25\textwidth]{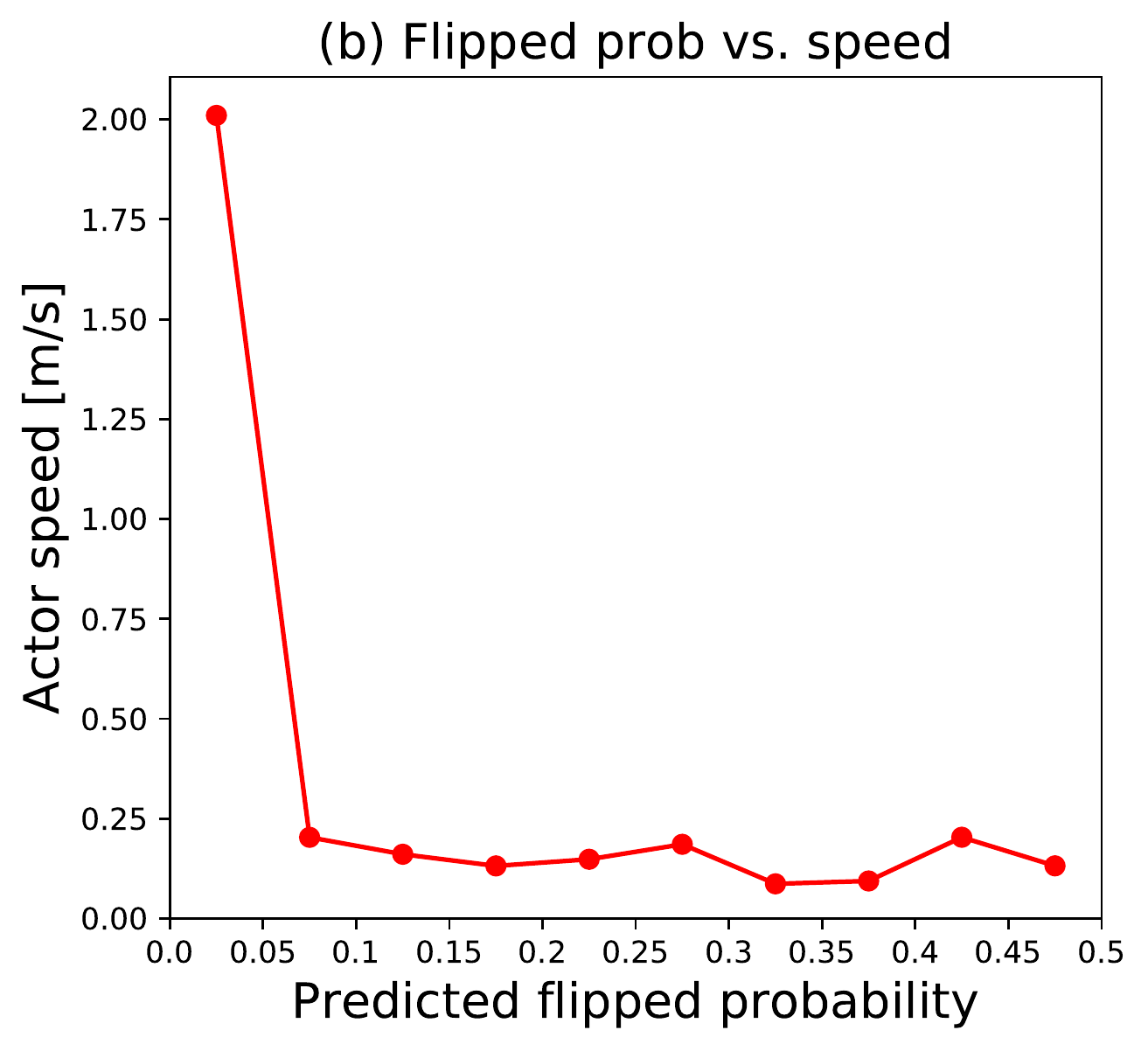}
    }
    \subfigure{
        \label{fig:flipped_prob_count}
        \includegraphics[width=0.25\textwidth]{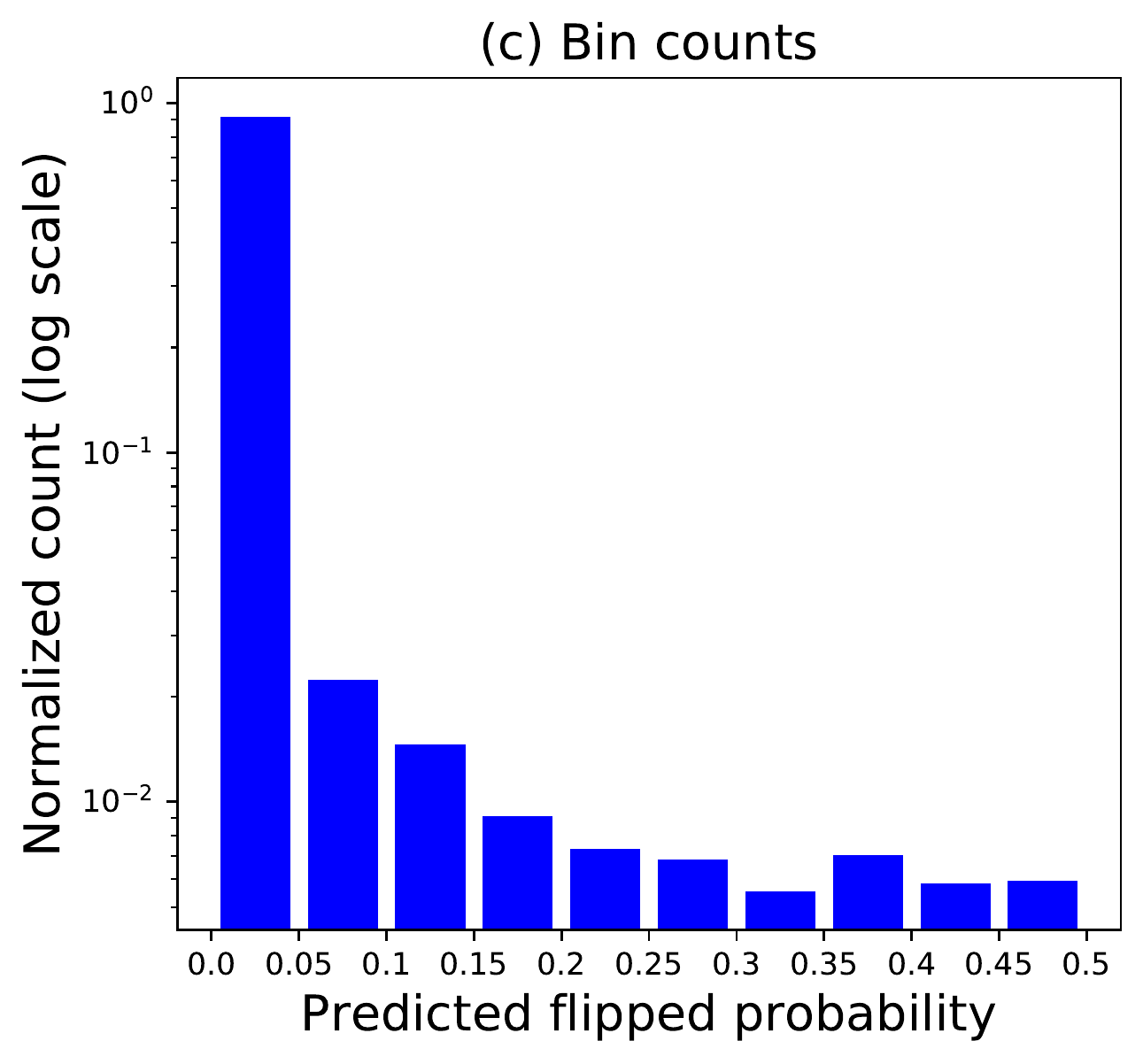}
    }
    \caption{Analysis of the orientation uncertainty outputs of the proposed \texttt{Flip-aware} method}
    \label{fig:flipped_prob}
    \vspace{-0.1in}
\end{figure*}

\section{Evaluation}

\subsection{Experimental setups}

{\bf Data set.}
We evaluated our method on the public nuScenes~\cite{caesar2020nuscenes} data set.
The data contains $1{,}000$ scenes collected from public roads in Boston and Singapore, sampled into $390{,}000$ frames at $20\text{Hz}$,
and we used the official partition for training and validation sets.

{\bf Implementation details.}
Our MultiXNet implementation used the same hyper-parameters as the original MultiXNet work~\cite{djuric2020multixnet}.
The rasterized BEV images have shape $L = 100\text{m}$, $W = 100\text{m}$, $V = 8\text{m}$, with resolution $\Delta_L = 0.125\text{m}$, $\Delta_W = 0.125\text{m}$, $\Delta_V = 0.2\text{m}$.
The model input is a history of $0.5\text{s}$ LiDAR sweeps ($T=10$ frames at $20\text{Hz}$) and the rasterized map, and the model outputs bounding boxes and predicts their trajectories for $3\text{s}$ into the future
($H = 30$ at $10\text{Hz}$).

{\bf Evaluation metrics.}
We used the standard KITTI~\cite{geiger2013vision} object detection and prediction metrics, including Average Orientation Similarity (AOS), Average Precision (AP), orientation errors, as well as trajectory $\ell_2$ errors.
While MultiXNet detects and predicts trajectories for all actor types, in this work we only focus only on the vehicle actors.
Nevertheless, we note that the metrics for other actor classes are mostly unchanged.
Following \cite{djuric2020multixnet} we used an IoU threshold of $0.7$ when computing the AP; note however that AP is IoU-based and a flipped bounding box will still be counted as a true positive.
For that reason, we also considered the AOS metric which weighs the precision by the average cosine distance of the orientations (normalized to a $[0, 1]$ range) at each recall point.
As a result, the AP metric is by definition an upper bound of AOS, and a completely flipped bounding box will have no positive contribution towards AOS.
We also measure the orientation error of the bounding boxes at $0\text{s}$
and the future trajectory $\ell_2$ error at $3\text{s}$
of the true positive detections, with the operating point set at $0.8$ recall using $0.5$ IoU threshold, as done in previous works~\cite{djuric2020multixnet, casas2020spagnn}.

For the orientation error we measure both full-range (FOE) and half-range orientation error (HOE), defined as
\begin{align}
\begin{split}
\label{eq:orientation_errors}
    \text{FOE} &= |(\theta_0 - \hat\theta_0) \mod 360\degree |, \\
    \text{HOE} &= |(\theta_0 - \hat\theta_0) \mod 180\degree |.
\end{split}
\end{align}
When evaluating the orientation and $\ell_2$ errors, we also slice the metrics by moving and non-moving actors.
We define an actor as moving if its ground-truth speed is larger than $0.5 \text{m/s}$, computed as a difference between the $0.5 \text{s}$ and $0 \text{s}$ ground-truth waypoints.
Lastly, for methods that output multiple orientations, such as our method and MultiBin~\cite{mousavian20173d}, we measured the errors of the highest-probability output.

{\bf Baselines.}
We compared our method against the other state-of-the-art methods listed in Table~\ref{tab:all-methods}, all added on top of the original MultiXNet losses.
For \texttt{MultiBin}~\cite{mousavian20173d} we used $n=2$ and $n=4$ bins, shown as \texttt{MultiBin-n} in the results. The bins are centered at $\{0\degree, 180\degree\}$ for $n=2$ and $\{-90\degree, 0\degree, 90\degree, 180\degree\}$ for $n=4$.
Lastly, to fairly evaluating FOE for \texttt{L1-sin} and \texttt{Sin-cos-2x}, we used a direction of the predicted trajectories to convert their half-range orientations to the full $360\degree$-range.

\subsection{Quantitative results}
\label{sec:quantitative_results}
The comparison of different approaches on the nuScenes data is given in Table~\ref{tab:main_results}, where the proposed method is denoted as \texttt{Flip-aware}.
We trained all models three times and report the means and standard deviations of the metrics, marking the best-performing methods in bold.

We can see that \texttt{Sin-cos-2x} had the best AP performance, and \texttt{L1-sin} had the best $\ell_2$ performance.
However, they were not able to represent full-range orientations as seen by large orientation errors.
Even when using the trajectory predictions to infer the orientations, their FOEs were still very high, especially when it comes to the non-moving actors.
As a result, they exhibited very low AOS.

When it comes to the full-range methods, \texttt{Sin-cos} had the lowest FOE.
However, it also had the worst AP performance among all the considered methods, trailing \texttt{Sin-cos-2x} by about 4\%.
This could be explained by the fact that the network incurs high losses when predicting a flipped bounding box, which causes the model to not spend enough learning capacity in accurately estimating the object detections.
The \texttt{MultiBin-n} models had similar performance compared to \texttt{Sin-cos}.
We hypothesize that MultiBin is not able to achieve better performance for our application as it was originally designed for 3D orientation estimation from camera images, while we estimate the object orientations from BEV voxels.

We can see that the proposed \texttt{Flip-aware} method achieved significant improvement in AP and HOE compared to the other full-range methods,
and it achieved the best AOS, AP, HOE, and moving FOE metrics among all the methods.
This demonstrates that our proposed loss design allows for very accurate full-range orientation estimation while achieving detection performance comparable to the best half-range methods.
In Section~\ref{sec:albation_study}, we investigate the reasons behind such performance, where we show that both the addition of the $\mathcal{L}_\text{half}$ loss term and the flipping-aware orientation estimation contributed to the improvements.
Moreover, our method is able to predict the probability that the orientation is actually flipped, helping the downstream modules to reason about scene uncertainties and allowing safer SDV operations.

\begin{table*} [t!]
\centering
\caption{Ablation study of the \texttt{Flip-aware} method}
\label{tab:ablation_study}
{\normalsize
{
  \begin{tabular}{lccccccc}
     &  &  & \multicolumn{3}{c}{\bf Orientation error [deg] $\downarrow$} & & \\
      \cmidrule(lr){4-6}
     & & & {\bf Half-range} & \multicolumn{2}{c}{\bf Full-range} & \multicolumn{2}{c}{\bf $\boldsymbol \ell_2$@$3\text{s}$ [m] $\downarrow$} \\
      \cmidrule(lr){5-6}\cmidrule(lr){7-8}
    {\bf Method} & {\bf $\text{AOS}_{0.7}$ $\uparrow$} & {\bf $\text{AP}_{0.7}$ $\uparrow$} & {\bf All} & {\bf All} & {\bf Moving} & {\bf All} & {\bf Moving} \\
    \hline
    \rowcolor{lightgray}
    \texttt{Flip-aware} & {\bf 57.9} & {\bf 60.7} & {\bf 1.71} & 9.6 & {\bf 2.2} & {\bf 0.99} & 2.75 \\
    \texttt{*-no-half} & 55.2 & 59.6 & 2.07 & 10.1 & 2.8 & 1.03 & 2.81 \\
    \rowcolor{lightgray}
    \texttt{*-no-flip} & 56.8 & 58.9 & 1.79 & {\bf 7.8} & 2.3 & {\bf 0.99} & {\bf 2.70} \\
    \hline
\end{tabular}
}
}
\end{table*}

\begin{figure*}[t!]
    \centering
    \subfigure{\includegraphics[trim={900 1000 750 900}, clip, width=0.40\textwidth]{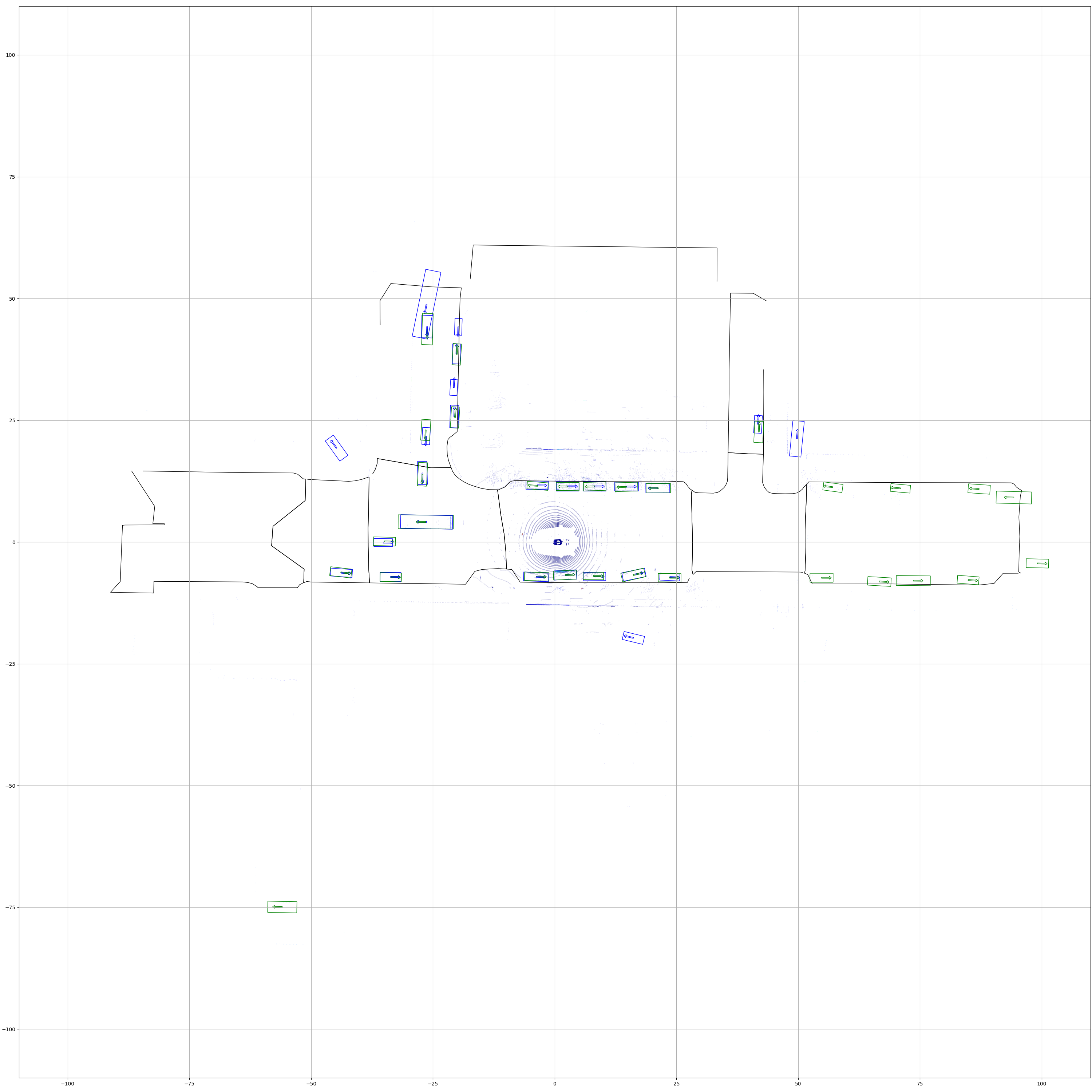}}
    \subfigure{\includegraphics[trim={900 1000 750 900}, clip, width=0.40\textwidth]{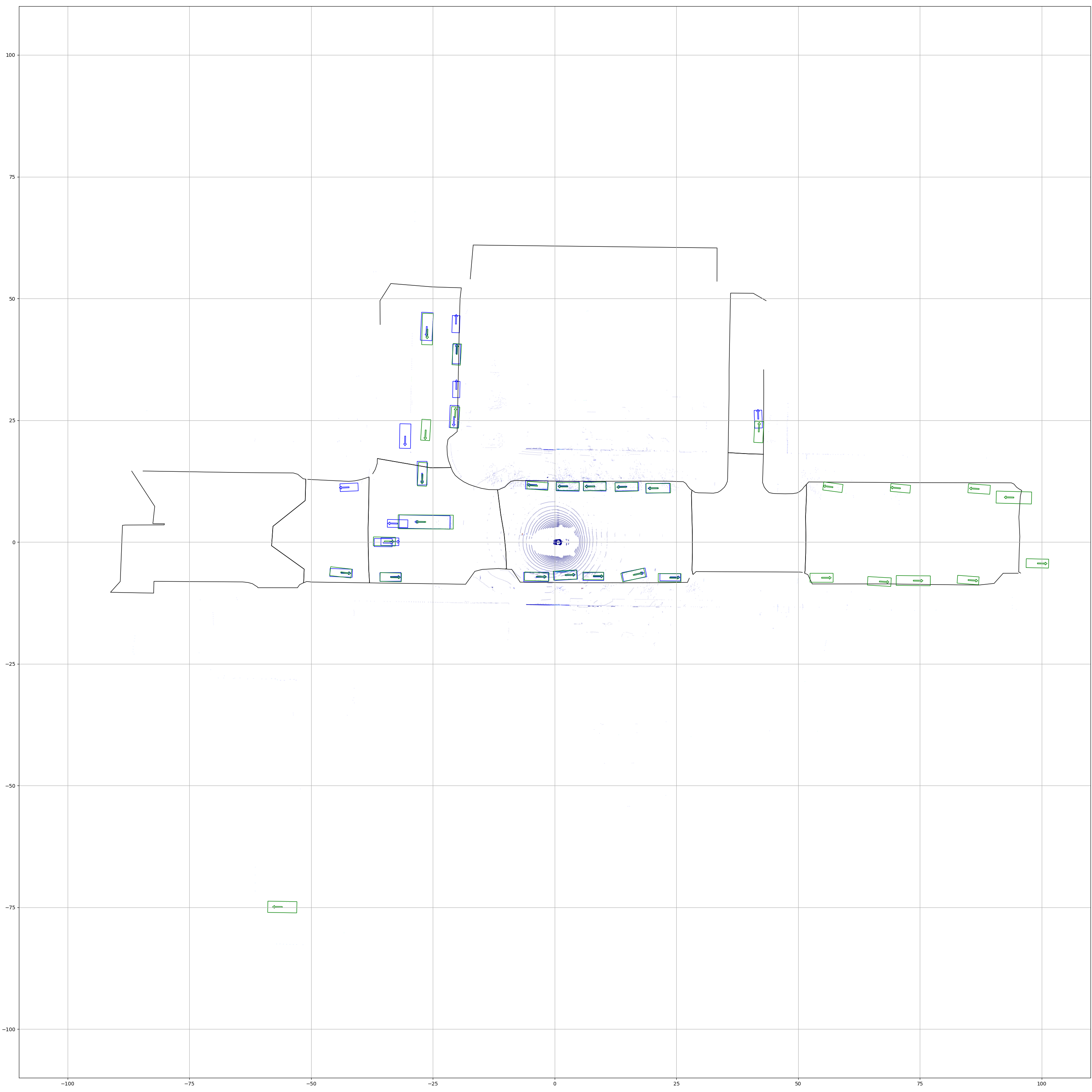}}
    \caption{Outputs of \texttt{Sin-cos-2x} and \texttt{Flip-aware}, respectively; ground truth shown in blue and outputs shown in green}
    \label{fig:qualitative_results}
\end{figure*}

\subsection{Analysis of uncertainty outputs}

In this section, we present the analysis of uncertainty outputs of the proposed \texttt{Flip-aware} method.
We binned all actors by their predicted flipped probabilities, and for each bin, we report the average FOE metric, as well as average speed and actor counts, illustrated in Figure~\ref{fig:flipped_prob}.
Note that, due to the post-processing step discussed in Section \ref{sect:half_full_combo}, the flipped probabilities of all actors are no greater than $0.5$.

Results in Figure~\ref{fig:flipped_prob}(a) show that when the model predicts an actor to have a low flipped probability, its FOE is also expected to be significantly lower than for those actors that have higher probabilities.
We can conclude that the predicted flipped probability is strongly correlated with the expected error, and is a reliable measure of our uncertainty in the actor's orientation.
In addition, in Figure~\ref{fig:flipped_prob}(b) and Figure~\ref{fig:flipped_prob}(c), we can see the average actor speed and the normalized actor count in each bin, respectively.
As expected, the model predicts lower flipped probabilities for moving actors than for non-moving actors, since the observed direction of motion is a strong indicator of an actor's orientation.
Moreover, for most of the actors, the model outputs low uncertainty, mirroring the distribution of the actor speeds in the data set.


\subsection{Ablation study}
\label{sec:albation_study}

To further understand the effectiveness of the two components of the proposed \texttt{Flip-aware} method, we performed an ablation study with two variants of the method, namely \texttt{Flip-aware-no-half} that does not include $\mathcal{L}_\text{half}$ loss term in equation \eqref{eq:flipping_aware_loss},
and \texttt{Flip-aware-no-flip} that does not model uncertainty and instead uses $\mathcal{L}_\text{full} + \mathcal{L}_\text{half}$ as its loss.
The experimental results are given in Table~\ref{tab:ablation_study}, showing that both models underperform on the AOS, AP, and HOE metrics, when compared to \texttt{Flip-aware}.
This demonstrates that both half-range loss and the uncertainty-aware component contributed to the detection performance improvements of our method. 
We can also see that \texttt{Flip-aware-no-flip} had better FOE performance, however, its AOS was lower, and importantly, the method does not provide the uncertainty outputs.

\subsection{Qualitative results}

In this section, we present the qualitative comparison of the two best performing half-range and full-range methods, namely \texttt{Sin-cos-2x} and \texttt{Flip-aware}.
As all methods performed reasonably well on moving actors, we focused on a scene with a large number of non-moving actors, shown in Figure \ref{fig:qualitative_results}. 
As shown in the example, the \texttt{Sin-cos-2x} method estimated incorrect orientations for a large number of parked vehicles.
This occurs because the method is trained to be indifferent to orientation flip through the design of its loss function, and moreover these parked vehicles have no moving trajectory predictions that could be used to reliably infer the orientations.
On the other hand, the proposed \texttt{Flip-aware} model predicted correct orientations for all vehicles in the scene, which is a result consistent with the previously seen strong performance in the quantitative evaluation.
We emphasize that correct orientation prediction for such parked and slow-moving vehicles is an important task for autonomous driving, leading to better behavioral modeling and improved safety during SDV operations.

\section{Conclusion}
\label{sect:conclusion}

In this paper we considered the problem of object detection and motion prediction in the context of self-driving technology. 
Unlike earlier work, we focused on the subtask of orientation prediction for vehicle actors. 
This is critical for full understanding of the actors' state and their motion prediction, especially for slow-moving and stopped vehicles. 
In addition to improved orientation prediction, the proposed approach also quantifies the prediction uncertainty by inferring the flipped probability, which is a very useful information for downstream modules in a self-driving system. 
Experiments on the real-world, open-source nuScenes data set indicate the benefits of the proposed method.

\balance

\newpage



\bibliographystyle{IEEEtran}
\bibliography{references}

\end{document}